%% file: main.tex
\documentclass{article}
\usepackage{ijcai19}
\usepackage[ruled,linesnumbered]{algorithm2e}
\SetAlCapNameFnt{\footnotesize}
\SetAlCapFnt{\footnotesize}
\usepackage{times}
\usepackage{soul}
\usepackage{url}
\usepackage[hidelinks]{hyperref}
\usepackage[utf8]{inputenc}
\usepackage[small]{caption}
\usepackage{graphicx}
\usepackage{amsmath}
\usepackage{booktabs}
\renewcommand\footnotemark{}
\usepackage{xcolor}

\DeclareMathOperator*{\argmin}{arg\,min}

 \definecolor{Sijia_color}{rgb}{0.858, 0.188, 0.478}
 \definecolor{xiaolong}{RGB}{181, 97, 2}

\DeclareMathAlphabet\mathbfcal{OMS}{cmsy}{b}{n}
\newcommand{\Def}[0]{\mathrel{\mathop:}=}

\title{SS-Auto: A Single-Shot, Automatic Structured Weight Pruning Framework of DNNs with Ultra-High Efficiency}

\author{
Zhengang Li\text{$^\dagger$}$^1$\and
Yifan Gong\text{$^\dagger$}$^1$\thanks{\hspace{-6.0mm}$^\dagger$These authors contributed equally.}\and
Xiaolong Ma$^{1}$\and
Sijia Liu$^{2}$\and
Mengshu Sun$^1$\and
Zheng Zhan$^{1}$\and \\
Zhenglun Kong$^{1}$\and
Geng Yuan$^{1}$\and
Yanzhi Wang$^{1}$
\affiliations
$^1$Northeastern University, 
$^2$IBM Watson Lab
\emails
$^1$\{li.zhen, gong.yifa, ma.xiaol, sun.meng, zhan.zhe, kong.zhe, yuan.geng\}@husky.neu.edu, yanz.wang@northeastern.edu, $^2$sijia.liu@ibm.com
}

\begin{document}

\maketitle

\begin{abstract}
Structured weight pruning is a representative model compression technique of DNNs for hardware efficiency and inference accelerations. Previous works in this area leave great space for improvement since sparse structures with combinations of different structured pruning schemes are not exploited fully and efficiently. To mitigate the limitations, we propose SS-Auto, a single-shot, automatic structured pruning framework that can achieve row pruning and column pruning simultaneously. We adopt soft constraint-based formulation to alleviate the strong non-convexity of $\ell_0$-norm constraints used in state-of-the-art ADMM-based methods for faster convergence and fewer hyperparameters. Instead of solving the problem directly, a Primal-Proximal solution is proposed to avoid the pitfall of penalizing all weights equally, thereby enhancing the accuracy. Extensive experiments on CIFAR-10 and CIFAR-100 datasets demonstrate that the proposed framework can achieve ultra-high pruning rates while maintaining accuracy. Furthermore, significant inference speedup has been observed from the proposed framework through actual measurements on the smartphone.

\end{abstract}

\input{introduaction.tex}

\input{background.tex}

\input{motivation.tex}

\section{Proposed SS-Auto Problem Formulation and Primal-Proximal Solution}\label{sec:scp}

The proposed SS-Auto problem formulation is, for the first time, a framework of the simultaneous row and column pruning for DNN models. We adopt the soft constraint-based formulation to alleviate the strong non-convexity of $\ell_0$-norm hard constraint in state-of-the-art ADMM formulation, thereby accelerating convergence and reducing the number of associated hyperparameters. We propose a Primal-Proximal solution to the problem to avoid the pitfall of penalizing all weights in the direct solution, resulting in accuracy enhancement. 
The single-shot, automatic nature of the proposed framework is attributed to the joint contribution of soft constraint-based formulation and proposed solution -- the former providing the possibility through hyperparameter reduction while the latter seamlessly integrating automatic per-layer pruning rate (or sparse distribution) determination in the solution to the Proximal problem.

\subsection{Problem Formulation using Soft Constraints}
We consider the pruning of a DNN model with $N$ layers. The weight collection $\mathbf W$ represents all GEMM weight matrices in CONV layers. The loss function of the DNN is denoted by $f(\mathbf W; \mathcal D)$, where $\mathcal{D}$ is the training dataset. We use the subscript to denote the layer index, rendering $\mathbf W = \{ \mathbf W_i \}_{i=1}^N$. 

The size of the weight matrix $\mathbf{W}_i$ for layer $i$ is $r_i\times c_i$. Let $[\mathbf W_{i}]_{p,:}$ and $[\mathbf W_{i}]_{:,q}$ denote the $p$-th row and the $q$-th column of $\mathbf W_{i}$, respectively. Here the bold letter will be reserved to represent a matrix or vector. Our objective is to achieve row pruning and column pruning simultaneously on the weight collection $\mathbf W$ while preserving the test accuracy. Therefore, the structured pruning problem can be formulated as an optimization problem:
\begin{align}\label{eq: prob}
    \begin{array}{cl}
\displaystyle  \min_{\mathbf W} \;
 f(\mathbf W; \mathcal D) +  \lambda  \sum_{i=1}^N  \bigg(
\sum_{p=1}^{r_i} \| [\mathbf W_{i}]_{p,:}  \|_2  \displaystyle + \sum_{q=1}^{c_i} \| [\mathbf W_{i}]_{:,q}  \|_2 \bigg ), 
    \end{array}
\end{align}
where $\sum_{p=1}^{r_i} \| [\mathbf W_{i}]_{p,:}  \|_2 + 
\sum_{q=1}^{c_i} \| [\mathbf W_{i}]_{:,q}  \|_2$
penalizes the row-wise and column-wise sparsity in 
$\mathbf W_{i}$, and the regularization parameter $\lambda > 0$ strikes a balance between the loss and the row/column-wise sparsity. 

Compared to the state-of-the-art ADMM-based pruning solution \cite{zhang2018systematic,zhang2018adam,ren2019ADMMNN} that relies on the strongly non-convex, $\ell_0$-norm based hard constraints, the soft constraint-based problem formulation requires much fewer hyperparameters (single $\lambda$ and $\rho$ versus the pre-defined per-layer pruning rate $\lambda$'s and convergence parameter $\rho$'s). There are threefold benefits: (\romannumeral1) accelerating convergence thanks to reduced non-convexity (reflected in the reduction in epoch number during training/pruning), (\romannumeral2) reducing the effort of hyperparameter determination in deriving a desirable solution, and (\romannumeral3) providing the possibility of automatic pruning rate determination with the help of explicit integration in the Proximal problem solution.

\subsection{Avoiding the Pitfall in Direct Solution}
A straightforward solution to the above problem is directly applying stochastic gradient descent using (\ref{eq: prob}) as the objective function. This direct solution is similar to the prior work using fixed regularizations \cite{wen2016learning,he2018amc}. However, a notable limitation of such a direct solution is that all weights in the DNN will be equally penalized. This is not aligned with the true objective in DNN weight pruning. As long as the target pruning rate is satisfied (or close), say 10$\times$ overall weight reduction, the remaining weights shall not be penalized anymore. Instead, the magnitudes of the remaining weights are likely to increase in order to compensate for the effect that a majority of weights are already pruned to zero. This divergence in objective function results in notable accuracy loss, and is the major limitation of the prior fixed regularization methods.

\vspace{-1mm}
\subsection{The Proposed Primal-Proximal Solution} \vspace{-1mm}
It is desirable to decouple the stochastic gradient descent process with regularization for the purpose of avoiding the pitfall in the direct solution. 
Through this decoupling and iterative process, we will enable a dynamic, gradual decrease of penalty on the remaining weights, which naturally fits with the desirable weight pruning process that more important weights should receive less penalty and be less likely to be removed. The penalty on weights will converge to zero in a systematic manner when the decoupled (Primal and Proximal) subproblems converge.
In order to decouple the process, we introduce \textit{two} auxiliary variables $\mathbf X$ and $\mathbf Y$, together with \textit{equality} constraints $\mathbf W = \mathbf X$ and $\mathbf W = \mathbf Y$, for row pruning and column pruning, respectively. Based on that, we reformulate problem \eqref{eq: prob} as
\begin{equation} \label{eq: prob1}
\begin{split}
    &\displaystyle  \min_{\mathbf W,\mathbf X,\mathbf Y} f(\mathbf W;\mathcal D)  + \lambda  \sum_{i=1}^N  \bigg(
\sum_{p=1}^{r_i} \| [\mathbf X_{i}]_{p,:}  \|_2  \displaystyle  + \sum_{q=1}^{c_i} \| [\mathbf Y_{i}]_{:,q}  \|_2 \bigg), \\
&\text{\quad s.t.\quad} \mathbf W = \mathbf X, ~ \mathbf W = \mathbf Y.
\end{split}
\end{equation}\vspace{-3mm}
The augmented Lagrangian function of \eqref{eq: prob1} is given by 
\begin{align}
    \mathcal L( \mathbf W, & \mathbf X, \mathbf Y, \boldsymbol{\Lambda }, \boldsymbol{\Gamma }) = f(\mathbf W; \mathcal D)  + \lambda  \sum_{i=1}^N  \bigg(
\sum_{p=1}^{r_i} \| [\mathbf X_{i}]_{p,:}  \|_2 \nonumber \\
& + \sum_{q=1}^{c_i} \| [\mathbf Y_{i}]_{:,q}  \|_2 \bigg) + \boldsymbol{\Lambda }^T ( \mathbf W - \mathbf X) 
+ \boldsymbol{\Gamma }^T ( \mathbf W - \mathbf Y) \nonumber \\
& + \frac{\rho}{2} \| \mathbf W - \mathbf X \|_2^2 
+ \frac{\rho}{2} \| \mathbf W - \mathbf Y \|_2^2,
\end{align}
where $ \boldsymbol{\Lambda }$ and $\boldsymbol{\Gamma}$
are Lagrangian multipliers, i.e., dual variables. $\rho > 0$ is a given augmented penalty value. For ease of notation, we view matrices as \textit{vectors} in optimization.

The Primal-Proximal solution (through augmented Lagrangian) is then given by the following alternating optimizations over two main subproblems (Primal) and (Proximal)\footnote{The Primal-Proximal decoupling process is analogous to ADMM framework in \cite{zhang2018systematic,zhang2018adam,ren2019ADMMNN}. However, the prior ADMM-based problem formulation is based on $\ell_0$ constraint; therefore, the second subproblem is simply a Euclidean mapping.}. At iteration $t$, the solution yields
\begin{align}
&\{ \mathbf W^{(t)}\} = \argmin_{\mathbf W} \; \mathcal L( \mathbf W, \mathbf X^{(t-1)},  \mathbf Y^{(t-1)}, \boldsymbol{\Lambda }^{(t-1)},  \boldsymbol{\Gamma }^{(t-1)})\tag{Primal} \label{eq: admm_primal} \\
&\{ \mathbf X^{(t)}, \mathbf Y^{(t)} \} = \argmin_{\mathbf X, \mathbf Y} \; \mathcal L( \mathbf W^{(t)}, \mathbf X, \mathbf Y,  \boldsymbol{\Lambda }^{(t-1)}, \boldsymbol{\Gamma }^{(t-1)})  & \tag{Proximal} \label{eq: admm_proximal} \\
& \boldsymbol{\Lambda}^{(t)} = \boldsymbol{\Lambda}^{(t-1)} + \rho ( \mathbf W^{(t)} - \mathbf X^{(t)} ) \nonumber \\
& \boldsymbol{\Gamma}^{(t)} = \boldsymbol{\Gamma}^{(t-1)} + \rho ( \mathbf W^{(t)} - \mathbf Y^{(t)}). \label{dual_update}
\end{align}
In each of the \eqref{eq: admm_primal} and \eqref{eq: admm_proximal} steps, $\mathcal{L}$ is minimized over the corresponding primal variables, using the most recent values of the other primal variables and dual variables. Step (\ref{dual_update}) is taken to update the dual variables. The dual variable $\mathbf{\Lambda}$ can be viewed as the scaled running sum of the consensus errors between $\mathbf{W}$ and $\mathbf{X}$. The same comprehension can be applied to $\mathbf{\Gamma}$ by replacing $\mathbf{X}$ with $\mathbf{Y}$. As the optimization process proceeds, $\mathbf{X}^{(t)}$ and $\mathbf{Y}^{(t)}$ converge to $\mathbf{W}^{(t)}$ gradually.

\paragraph{Solving \eqref{eq: admm_primal}-problem.}
Problem \eqref{eq: admm_primal} can be simplified as
\begin{align}\label{eq: prob_primal_1}
    \begin{array}{cl}
\displaystyle \min_{\mathbf W} \; f(\mathbf W; \mathcal D)  + \frac{\rho}{2} \| \mathbf W - \mathbf B_1 \|_2^2  + \frac{\rho}{2} \| \mathbf W - \mathbf B_2 \|_2^2 , 
    \end{array}
\end{align}
where $\mathbf B_1 \Def (\mathbf X^{(t-1)} - (1/\rho) \boldsymbol{\Lambda}^{(t-1)})$ and 
$\mathbf B_2 \Def (\mathbf Y^{(t-1)} - (1/\rho) \boldsymbol{\Gamma}^{(t-1)})$.
The first term in \eqref{eq: prob_primal_1} is the differentiable loss function while the second and the third terms are quadratic and differentiable. Thus, this subproblem can be solved by standard solvers efficiently.


\paragraph{Solving \eqref{eq: admm_proximal}-problem.}

Problem \eqref{eq: admm_proximal} can be equivalently decomposed over $\mathbf X$ and $\mathbf Y$, leading to problems
\begin{align}\label{eq: prob_X}
\begin{array}{cl}
\displaystyle \min_{\mathbf X} \; \sum_{i=1}^N  \bigg (
\sum_{p=1}^{r_i} \| [\mathbf X_{i}]_{p,:}  \|_2 
\bigg ) + \frac{\rho }{2\lambda} \| \mathbf X -  \mathbf C_1\|_2^2
\end{array}, \\
\begin{array}{cl}
\displaystyle \min_{\mathbf Y} \;
  \sum_{i=1}^N  \bigg (
\sum_{q=1}^{c_i} \| [\mathbf Y_{i}]_{:,q}  \|_2 
\bigg ) + \frac{\rho }{2\lambda} \| \mathbf Y -  \mathbf C_2\|_2^2,
\end{array} \label{eq: prob_Y}
\end{align}
where $\mathbf C_1 \Def (\mathbf W^{(t)} + (1/\rho) \boldsymbol{\Lambda}^{(t-1)})$ and $\mathbf C_2 \Def (\mathbf W^{(t)} + (1/\rho) \boldsymbol{\Gamma}^{(t-1)})$. Solving problem \eqref{eq: prob_X} corresponds to finding the proximal operator of $\sum_{i=1}^N  \bigg (\sum_{q=1}^{c_i} \| [\mathbf Y_{i}]_{:,q}  \|_2\bigg )$ with parameter $\lambda/\rho$, which compromises between minimizing the first term of \eqref{eq: prob_X} and being near to $\mathbf{C}_1$. A similar exposition can be obtained for problem \eqref{eq: prob_Y}. These two problems can be solved in parallel, thus achieving the goal of finding the row-wise and column-wise sparsity jointly. 

Analytical solutions can be found for problems \eqref{eq: prob_X} and \eqref{eq: prob_Y}. Based on \cite{parikh2014proximal}, the solution to problem \eqref{eq: prob_X} is given by
\begin{align} \label{prox_x}
    [\mathbf X_{i}^{(t)}]_{p,:} \! = \! \left \{ 
    \begin{array}{cl}
    \!\!\!(1 - \frac{\lambda}{\rho \| [\mathbf C_{1,i} ]_{p,:} \|_2})  [\mathbf C_{1,i} ]_{p,:} \!\!\!& \text{if } \| [\mathbf C_{1,i} ]_{p,:} \|_2 \! \geq \! \lambda/\rho \\
    \mathbf 0 & \text{otherwise}
    \end{array}
    \right.
\end{align}
where $\mathbf C_{1,i} $ is the submatrix of $\mathbf C_1$ corresponding to $\mathbf X_{i}$.
Similarly, the solution to problem \eqref{eq: prob_Y} is given by
\begin{align} \label{prox_y}
    [\mathbf Y_{i}^{(t)}]_{p,:} \! = \! \left \{ 
    \begin{array}{cl}
    \!\!\!(1 - \frac{\lambda}{\rho \| [\mathbf C_{2,i} ]_{p,:} \|_2})  [\mathbf C_{2,i} ]_{p,:} \!\!\!& \text{if } \| [\mathbf C_{2,i} ]_{p,:} \|_2 \! \geq \! \lambda/\rho \\
    \mathbf 0 & \text{otherwise}
    \end{array}
    \right.
\end{align}
where $\mathbf C_{2,i} $ is the submatrix of $\mathbf C_2$ corresponding to $\mathbf Y_{i}$.
\subsection{Summary of Pruning Algorithm}
The main steps of our structured pruning method are demonstrated in Algorithm \ref{alg:prox}. Given a pretrained DNN model with parameters $\mathbf{W}^{(0)}$, our method iterates $T$ steps to find the sparse structure. During the training process, we incorporate the penalty for row-wise and column-wise sparsity into the objective function. At each iteration $t$, the algorithm first updates the weight $\mathbf{W}^{(t)}$ by solving problem (\ref{eq: admm_primal}) using stochastic gradient descent. Then, it proceeds to solve problem (\ref{eq: admm_proximal}) according to (\ref{prox_x}) and (\ref{prox_y}). The dual variables $\mathbf{\Lambda}^{(t)}$ and $\mathbf{\Gamma}^{(t)}$ are then calculated with the newly updated primal variables. After training, the weights that are close to zero are removed. A retraining phase is taken to update the remaining weights to restore accuracy.

\vspace{-1.5mm}
\begin{algorithm}[htp]
\footnotesize
\caption{Overall Primal-Proximal Pruning Algorithm}
\label{alg:prox}
\SetKwInOut{Input}{Input}\SetKwInOut{Output}{Output}
\Input{Pretrained DNN model $\mathbf{W}^{(0)}$, total iteration $T$; regularization parameter $\lambda$; augmented penalty $\rho$}
\Output{Structured pruned model with both row-wise and column-wise sparsity}
Initialize $\mathbf X^{(0)}\leftarrow\mathbf W^{(0)}$, $\mathbf Y^{(0)}\leftarrow\mathbf W^{(0)}$, $\mathbf{\Lambda}^{(0)}\leftarrow\bf{0}$, $\mathbf{\Gamma}^{(0)}\leftarrow\bf{0}$ \;
\For{\rm{iteration} $t \leftarrow1$ \rm{to} $T$}{
Update $\mathbf{W}^{(t)}$ by solving (\ref{eq: admm_primal}) with standard solvers\;
Calculate $\mathbf{X}^{(t)}$ and $\mathbf{Y}^{(t)}$ according to (\ref{prox_x}) and (\ref{prox_y}) to obtain row-wise and column-wise sparse distributions\;
Update dual variables $\mathbf{\Lambda}^{(t)}$ and $\mathbf{\Gamma}^{(t)}$ according to (\ref{dual_update})\;
}
Remove the group of weights according to the sparse distributions in $\mathbf X^{(T)}$ and $\mathbf Y^{(T)}$ and retrain the rest of non-zero weights to retain accuracy\;
\end{algorithm}
\vspace{-1.5mm}
Our Primal-Proximal pruning algorithm has the following favorable features. First, the algorithm has no requirement for the pre-defined per-layer pruning rate and contains much fewer hyperparameters, only the regularization parameter $\lambda$ and the augmented penalty $\rho$, compared with state-of-the-art ADMM-based algorithms. Second, the pruning rate, as well as the row(s) and column(s) to be pruned that minimize the loss value while maximizing the reduction in the storage and computation can be obtained automatically via solving Problem (\ref{eq: admm_proximal}) in each iteration during training. Combining these two features, we conclude that the Primal-Proximal pruning algorithm of SS-Auto has the advantage of finding the per-layer pruning rate automatically in one shot.











\input{experiment_results.tex}

\section{Conclusion}
We present SS-Auto, a single-shot, automatic structured weight pruning framework of DNNs that can find row-wise and column-wise sparsity simultaneously. A soft constraint-based formulation is adopted to alleviate the strong non-convexity of $\ell_0$-norm based hard constraints in state-of-the-art ADMM formulation. A Primal-Proximal solution is proposed to address the structured pruning problem for accuracy enhancement. Extensive experiments on CIFAR-10 and CIFAR-100 datasets demonstrate that Auto-SS can achieve ultra-high compression rate and fast inference while preserving accuracy. 


\end{document}

%% file: introduaction.tex
\section{Introduction}

Deep neural networks (DNNs) have achieved human-level performances in various application domains including image classification \cite{krizhevsky2012imagenet}, speech recognition \cite{hinton2012deep}, and natural language processing \cite{Wu2016GooglesNM}. However, the ever-increasing parameters and computations of DNNs~\cite{simonyan2014very,he2016deep} bring difficulties for DNNs to execute in storage and power budget limited situations, especially, the mobile devices. To mitigate the challenge, model compression techniques for DNNs~\cite{han2015learning,wen2016learning,min20182pfpce,he2018amc,zhang2018systematic,zhang2018adam} have been intensively studied to simultaneously reduce the model size and accelerate inference computations, thereby meeting the requirements of practical applications. 

As one representative model compression technique, weight pruning explores the redundancy in the number of weights of DNN models. Early works of non-structured weight pruning~\cite{han2015learning,guo2016dynamic,liu2018rethinking} leverage various heuristics to prune less important weights at arbitrary locations. Despite the high compression rate, non-structured pruning methods are not friendly to inference accelerations on hardware, because the remaining weights are distributed in irregular, sparse matrices which demand additional indices for storage/computation. On the other hand, structured pruning methods \cite{min20182pfpce,Zhuang2018DCP,zhu2018ijcai,ma2019tiny,Zhao2019VariationalCN,Liu2020Autocompress} directly reduce the size of the weight matrix while maintaining a full matrix format, eliminating the storage requirements for indices. Therefore, structured pruning is more compatible with hardware accelerations and has become the main research focus. 

Recently, research works \cite{zhang2018systematic,zhang2018adam,ren2019ADMMNN} achieve substantial weight reduction and preserve promising accuracy with the successful applications of the powerful Alternating Direction Methods of Multipliers (ADMM) framework. ADMM outperforms previous direct approaches by avoiding the equal penalty on all weights. The solution framework is potent and applicable to different schemes of structured pruning and non-structured pruning. However, previous ADMM-based approaches reach the compression target through setting $\ell_0$-norm based hard constraints that contain a large number of hyperparameters. The strong non-convexity of $\ell_0$-norm causes a long convergence time to obtain the sparse structure. Furthermore, it is highly time-consuming to manually explore the vast design space of hyperparameters trading off among model size, speed, and accuracy, and the derived hyperparameters are typically sub-optimal. An alternative way is to employ an automated process \cite{he2018amc,Liu2020Autocompress} of hyperparameter determination for such structured pruning problems. But these methods are not one-shot solutions and involve an iterative process to find the appropriate hyperparameters, resulting in a significant increase in the overall training time. Nevertheless, none of these existing approaches can find effective combinations of different structured pruning schemes simultaneously for a higher degree of weight reduction and accuracy maintenance. 

To address the limitations of prior works, we propose SS-Auto, a single-shot, automatic structured pruning framework that can achieve row pruning and column pruning simultaneously. We begin by adopting the soft constraint-based formulation with penalty terms for both row-wise and column-wise sparsity. Through avoiding the strongly non-convex $\ell_0$-norm based hard constraints in the state-of-the-art ADMM formulation, convergence can be accelerated, and the number of associated hyperparameters can be reduced. Furthermore, the formulation provides the possibility of deriving the per-layer pruning rate automatically. Instead of solving the problem directly, a Primal-Proximal solution is introduced to avert penalizing all weights equally by decoupling the stochastic gradient descent process with regularization. The decoupling and iterative process enables a progressive, gradual decrease of the penalty on the remaining weights, which naturally fits with the desirable pruning process that more important weights should receive less penalty and be less inclined to be pruned, thus increasing the accuracy. Meanwhile, the solution of the Proximal problem renders an automatic determination of the per-layer pruning rate.

Combining all the improvements results in our framework SS-Auto, which surpasses prior works by up to 20.5$\times$ in weight reduction. Extensive experiments on CIFAR-10 and CIFAR-100 prove the effectiveness of SS-Auto in finding sparse structures through automatic pruning rate determination and joint row and column pruning. Furthermore, significant inference speedup has been observed from the SS-Auto framework through actual measurements on mobile with our compiler-assisted mobile DNN acceleration framework. 

%% file: background.tex
\section{Background on DNN Weight Pruning}
Weight pruning methods remove redundant or less important weights in DNNs to reduce the storage and computational costs for the inference phase. There exist two mainstreams of weight pruning, i.e., non-structured pruning and structured pruning. 

\textbf{Non-structured pruning:} Non-structured weight pruning is fine-grained and prunes weights at arbitrary locations. The early work \cite{han2015learning} leverages a heuristic method to prune weights with small magnitudes iteratively. With the successful applications of the powerful ADMM optimization framework, research works \cite{zhang2018systematic,ren2019ADMMNN} achieve a very high weight reduction rate while maintaining promising accuracy. However, non-structured methods lead to sparse and irregular weight matrices, which require indices to be stored in a compressed format. A representative sparse representation format is the compressed sparse row (CSR), adopted in \cite{han2015learning,han2016eie}. Though saving the storage cost, the decoding of each stored index requires a search over the whole activation vector. Consequently, it suffers from limited accelerations in actual hardware implementations. 

\begin{figure}[t]
    \centering
    \includegraphics[width=0.45 \textwidth]{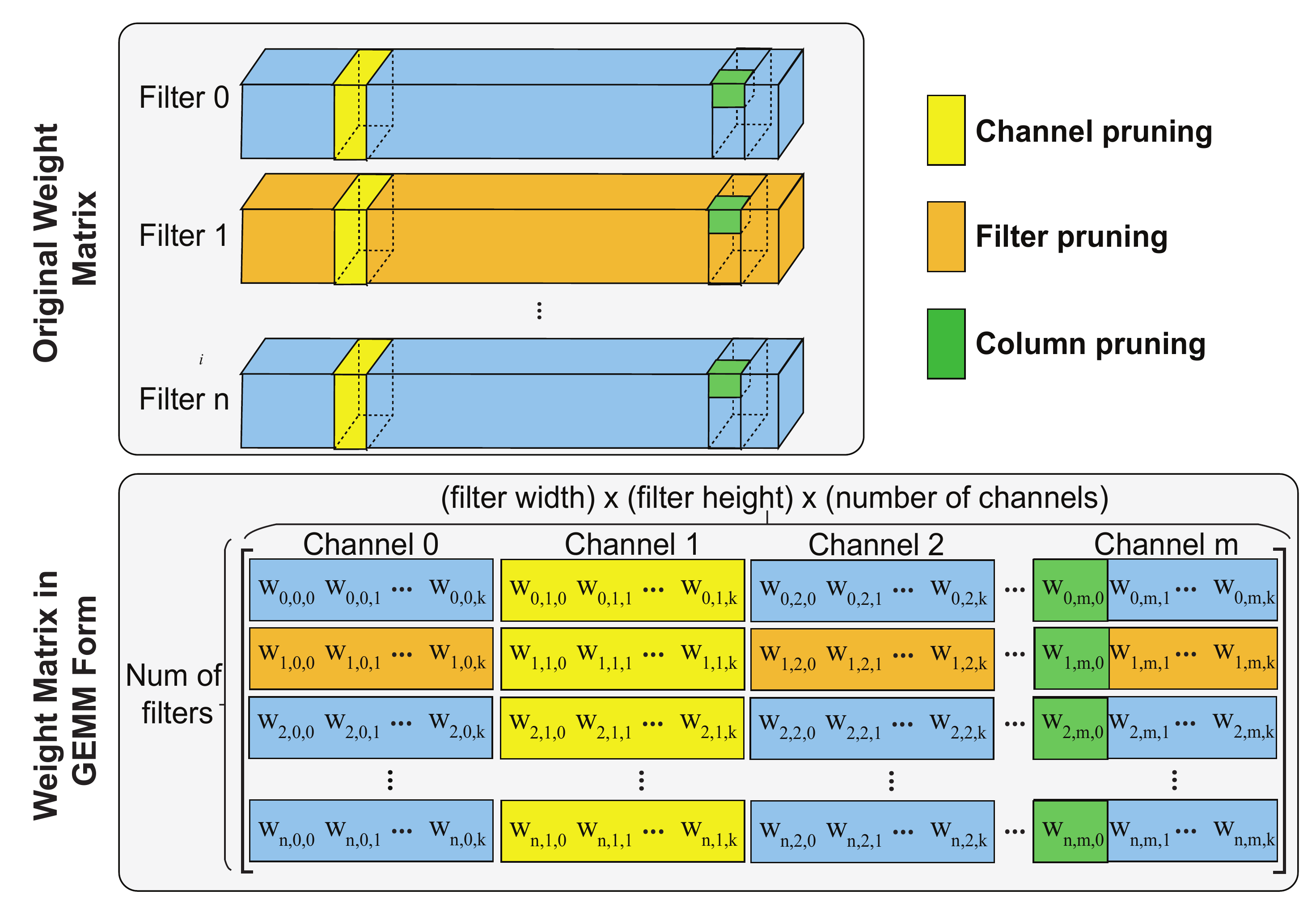}
    \vspace{-0.2mm}
    \caption{{\bf Representative structured weight pruning schemes.}}
    \label{fig:GEMM}
    \vspace{-5mm}
\end{figure}

\textbf{Structured pruning:} To overcome the limitations of non-structured pruning, recent works \cite{li2016pruning,min20182pfpce,Zhuang2018DCP,zhu2018ijcai,he2018amc,ma2019tiny,Zhao2019VariationalCN,Liu2020Autocompress} proposed to incorporate regularity in weight pruning with a main focus on convolution (CONV) layers of DNNs. Figure \ref{fig:GEMM} illustrates three types of structured pruning, i.e., filter pruning, channel pruning, and filter shape pruning. Filter pruning removes the entire filter(s), channel pruning removes whole channel(s), and filter shape pruning removes the weights in the same location(s) of all filters in a layer. Convolutional computation in DNNs is commonly transformed into general matrix multiplication (GEMM) by converting weight tensors and feature map tensors to matrices. Filter pruning is also termed row pruning since it corresponds to reducing one row of the weight matrix, as shown in Figure \ref{fig:GEMM}. Accordingly, filter shape pruning is also termed column pruning. Furthermore, channel pruning corresponds to reducing multiple consecutive columns. These three structured pruning schemes, along with their combinations, reduce the dimensions in GEMM while maintaining a full matrix format, thereby facilitating hardware implementations. 

The major advantage of filter/channel pruning is the superlinear effect on storage/computation reduction. Namely, $\alpha\times$ filter/channel pruning on all layers results in over $\alpha\times$ reduction in the number of weight parameters. As for column pruning, it has a higher degree of flexibility. These schemes can be combined for a higher reduction rate on the computation and storage. In this way, an effective method for finding a desirable combination is needed.

%% file: motivation.tex
\section{Solution Techniques: Prior Work and Motivations}
 Though non-structured weight pruning has the advantage of a high compression rate, recent research works \cite{li2016pruning,min20182pfpce,Zhuang2018DCP,Wang2019NonstructuredDW} have proven that the irregular, sparse structures caused by non-structured pruning are typically not compatible with GPU or multi-core CPU accelerations for DNN inference. Therefore, we focus on structured pruning techniques within this paper. 

\textbf{ADMM:} As an effective mathematical optimization technique, \emph{Alternating Direction Method of Multipliers} (ADMM) deals with non-convex optimization problems by decomposing the original problem into two subproblems that can be solved separately and efficiently \cite{boyd2011distributed}. Compared with prior works using fixed regularization methods for the pruning problem \cite{wen2016learning,he2018amc}, ADMM avoids penalizing all the weights equally by solving the subproblems iteratively, thereby yielding improvements in the quality of results. 

To leverage the ADMM optimization technique, current weight pruning approaches \cite{zhang2018systematic,zhang2018adam,ren2019ADMMNN} formulate the pruning problem as an optimization problem with $\ell_0$-norm based hard constraints specifying the sparsity requirements. The associated constraints for both structured pruning \cite{zhang2018adam} and non-structured pruning \cite{zhang2018systematic,ren2019ADMMNN} belong to the subset of combinatorial constraints that can be solved by ADMM \cite{hong2016convergence}. For instance, the $\ell_0$-norm hard constraint for row-wise structured pruning restricts the number of nonzero filters in a layer to be smaller than a predefined value. The value is pivotal as it determines the pruning rate and influences accuracy. Hence, an appropriate selection of the per-layer pruning rate is required for the successful application of ADMM to the weight pruning problem. However, manually determining the pruning rate for each layer involves a highly time-consuming trial-and-error process, and the derived hyperparameters are typically sub-optimal. Furthermore, $\ell_0$-norm based hard constraints are strongly non-convex, leading to long convergence time for the acquisition of the pruning results. 

Besides, a proper combination of row pruning and column pruning can yield better performance compared with applying either of these two schemes solely. Previous methods using ADMM to find the combinations address the row pruning problem and column pruning problem one by one separately. Nevertheless, a better approach is to solve these two problems jointly, considering the influence between each other.

\textbf{AutoML:} Inspired by the concept of automated machine learning (AutoML) \cite{zoph2017NAS,real2017ICML}, recent works propose to employ an automated process of hyperparameter determination for the structured pruning problem, instead of exploring the hyperparameters manually. AMC \cite{he2018amc} leverages deep reinforcement learning to generate the pruning rate for each layer of a target model. However, it only considers the row pruning scheme and adopts an early weight pruning technique based on fixed regularization. AutoCompress \cite{Liu2020Autocompress} makes further improvements in this direction by incorporating the combination of structured pruning schemes into the search space and adopting ADMM-based structured weight pruning as the core algorithm. Moreover, AutoCompress replaces the prior deep reinforcement learning technique, which has an underlying incompatibility with the weight pruning problem, with a heuristic search method enhanced by an experience-based guided search. Such automated methods reduce the burdensome manual feature selection process, but they are not one-shot solutions and require many rounds to decide the hyperparameters, resulting in a significant increase in the overall training time.    

Motivated by the limitations of state-of-the-art approaches, we aim to come up with a structured pruning method that possesses the following characteristics: (\romannumeral1) can solve the row pruning problem and column pruning problem jointly; (\romannumeral2) can significantly reduce the amount of hyperparameters to be determined; (\romannumeral3) can provide the possibility of determining the pruning rates automatically in a single shot.

%% file: experiment_results.tex
\section{Experimental Results}\label{sec:acc_results}

\subsection{Experiment Settings}
To evaluate the effectiveness of SS-Auto, we conduct extensive experiments on three representative network structures, i.e., VGG-16, ResNet-18, and MobileNet-V2, with two major image classification datasets, i.e., CIFAR-10 and CIFAR-100. We focus on the pruning of CONV layers, which are the most computationally intensive layers in modern DNNs. The implementations are based on Pytorch. We use an augmented penalty $\rho=10^{-3}$, regularization parameter $\lambda=10^{-7}$, and learning rate $10^{-4}$. The number of iterations $T$ is set as 300 (one iteration per epoch). The ADAM optimizer is utilized. 

To demonstrate the actual inference performance of SS-Auto on mobile platforms, we show our results on a mobile device, based on our compiler-assisted mobile DNN acceleration framework. The compiler-assisted platform is used to guarantee end-to-end inference execution efficiency by enabling optimized code generation. As the computation paradigm of DNN is in a manner of layerwise execution, we convert a DNN model into computational graph, embodied by static C++ (for CPU execution) or OpenCL (for GPU execution) code. The speed measurements are conducted on a Samsung Galaxy S10 cell phone with the latest Qualcomm Snapdragon 855 mobile platform that consists of a Qualcomm Kryo 485 Octacore CPU and a Qualcomm Adreno 640 GPU. 

\begin{figure}[t]
\vspace{-4mm}
    \centering
    \includegraphics[width=0.49 \textwidth]{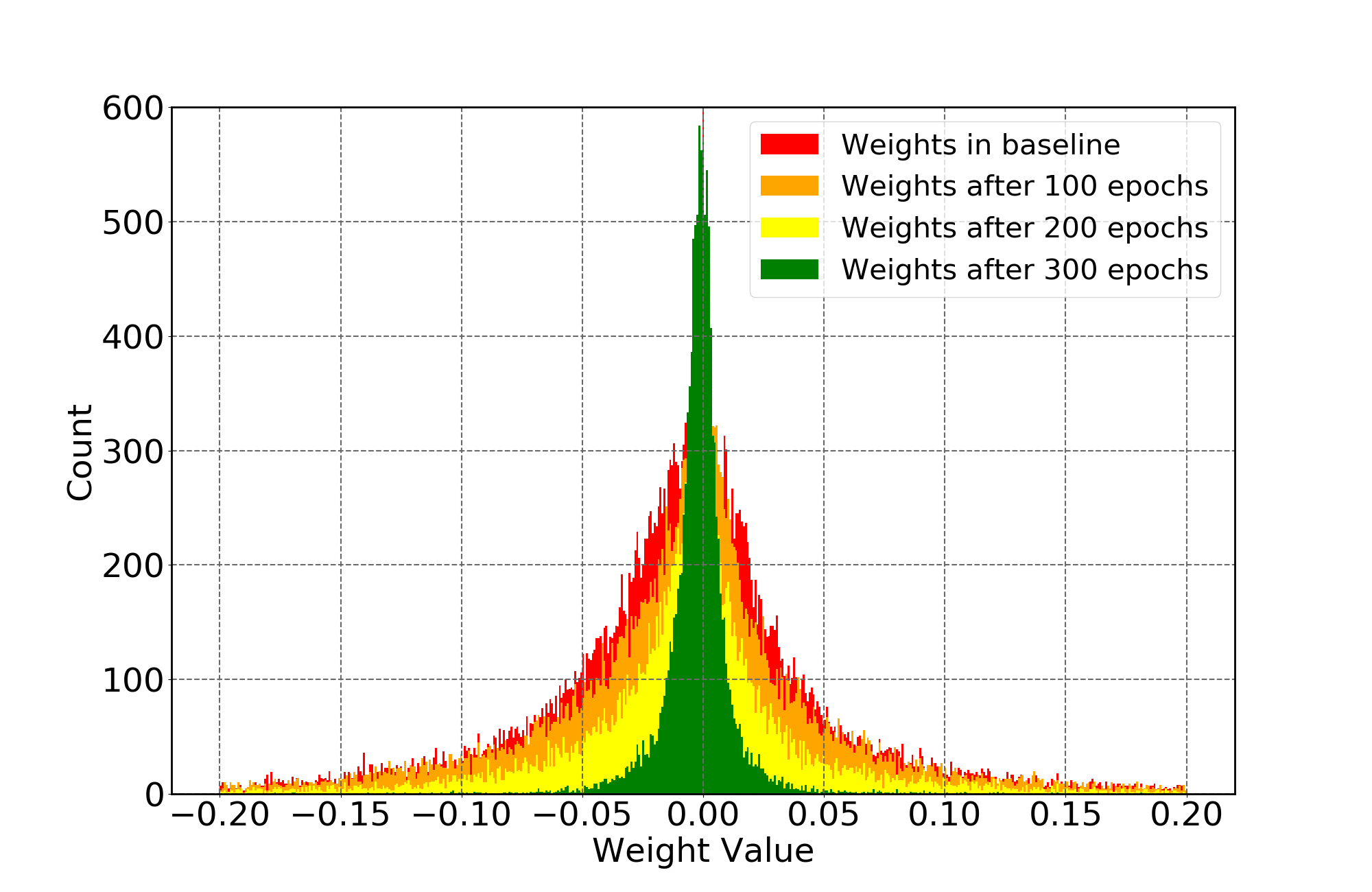}
    \vspace{-2mm}
    \caption{{\bf Weights distribution generated by SS-Auto in the second CONV layer of VGG-16 model with CIFAR-10 dataset.}}
    \label{fig:weight distribution}
    \vspace{-3mm}
\end{figure}

\begin{figure*}[t]
    \centering
    \includegraphics[width=0.95 \textwidth]{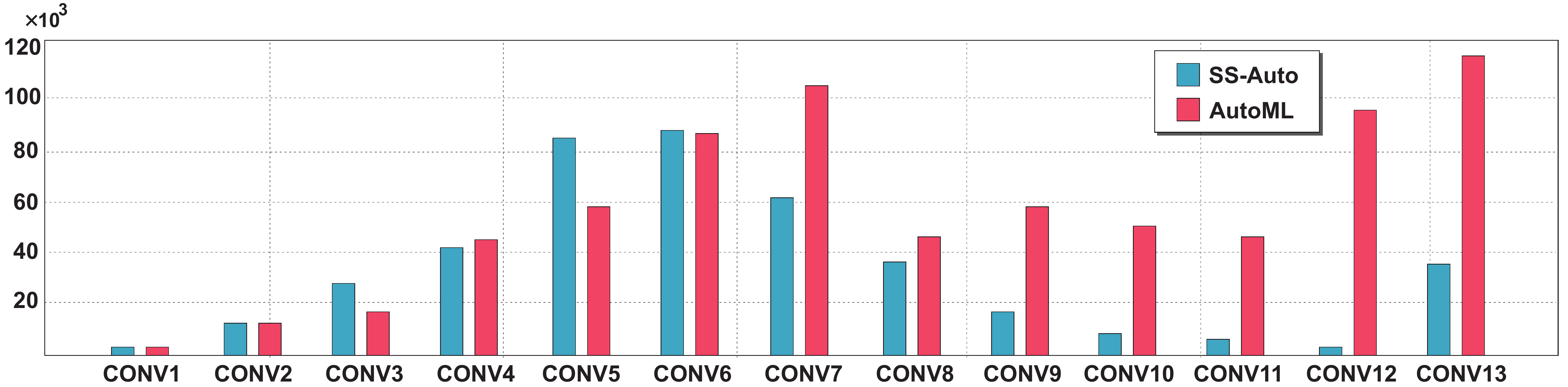}
    \vspace{-1mm}
    \caption{Remaining parameters counts in each CONV layer of VGG-16 (CIFAR-10). Our method (34.5 $\times$) vs. AutoCompress (22.7 $\times$).}
    \label{fig:weght in layer}
     \vspace{-4mm}
\end{figure*}

\subsection{Runtime Analysis}
The overall pruning algorithm of SS-Auto, as shown in Algorithm \ref{alg:prox}, mainly includes two parts: the Primal-Proximal pruning (or training) process and the retraining process. For the first part, we update the primal variables $\mathbf X$, $\mathbf Y$, and dual variables $\mathbf{\Lambda}$, $\mathbf{\Gamma}$ in each iteration, leading to the sparse structure gradually. Figure \ref{fig:weight distribution} shows the process of sparse distribution generation at the second CONV layer of VGG-16 on CIFAR-10 dataset. As can be seen from the figure, as the pruning process proceeds, less important weights are removed gradually with the help of the proposed Primal-Proximal solution.   

Despite an iterative process, the total number of epochs and training time of SS-Auto are limited. For instance, SS-Auto only requires 300 epochs for the Primal-Proximal pruning process and 300 epochs for retraining when pruning VGG-16 on the CIFAR-10 dataset. This is even fewer than the original ADMM-based solution with manually determined per-layer pruning rates, which requires 1,000-1,200 epochs for the pruning process. As for AutoCompress, it typically takes 2,000-3,000 epochs for the loss to converge during pruning due to the automatic search of the pruning rates. The improvement in convergence time is attributed to the adoption of the soft constraint-based formulation, which alleviates the strong non-convexity of $\ell_0$-based hard constraints in ADMM-based methods.




\begin{table}[!t]\scriptsize
\vspace{2pt}
\caption{\textbf{SS-Auto} vs. other pruning methods on CIFAR-10.} \label{Table1}
\vspace{-0.4mm}
\resizebox{0.48 \textwidth}{!}{
\label{table1}
\begin{tabular}{|cccccc|}
\hline
 Methods & \begin{tabular}[c]{@{}c@{}}Base \\ Accuracy\end{tabular} & \begin{tabular}[c]{@{}c@{}}Prune\\ Accuracy\end{tabular} & \begin{tabular}[c]{@{}c@{}}Conv\\ Comp.rate\end{tabular} & \begin{tabular}[c]{@{}c@{}}Sparsity\\ Type\end{tabular} & 
 \begin{tabular}[c]{@{}c@{}}Inference\\ Time (ms) \\CPU/GPU\end{tabular}\\ 
 \hline
 \hline
 \multicolumn{6}{|c|}{VGG-16} \\
 \hline
 Iter. Pruning~\shortcite{han2015learning} & 92.5\% & 92.2\% & 2.0$\times$ & Irregular & N/A \\
 One Shot~\shortcite{liu2018rethinking} & 92.5\% & 92.4\% & 2.5$\times$ & Irregular & N/A\\
 2PFPCE~\shortcite{min20182pfpce} & 92.9\% & 92.8\% & 4.0$\times$ & Structured & N/A\\
 Effi. ConvNet~\shortcite{li2016pruning} & 93.2\% & 93.4\% & 2.7$\times$ & Structured & N/A\\
 ADMM~\shortcite{ma2019tiny} & 93.7\% & 93.4\% & 20.2$\times$ & Structured & N/A\\
 AutoCompress~\shortcite{Liu2020Autocompress} & 93.7\% & 93.6\% & 22.7$\times$ & Structured & 2.91/2.74 \\ 
 \textbf{SS-Auto} & 93.7\% & 93.7\% & 34.4$\times$ & Structured & 2.68/2.47 \\ 
 \textbf{SS-Auto} & 93.7\% & 93.5\% & 41.1$\times$ & Structured & 2.52/2.32 \\ 
 \hline
 \hline
 \multicolumn{6}{|c|}{ResNet-18} \\
 \hline
 DCP~\shortcite{Zhuang2018DCP} & 88.9\% & 87.6\% & 2.0$\times$ & Structured & N/A \\
 AMC~\shortcite{he2018amc} & 90.5\% & 90.2\% & 2.0$\times$ & Structured & N/A\\
 Vari. Pruning~\shortcite{Zhao2019VariationalCN} & 92.0\% & 91.7\% & 1.6$\times$ & Structured & N/A\\
ADMM~\shortcite{ma2019tiny} & 94.1\% & 93.2\% & 15.1$\times$ & Structured & N/A \\
AutoCompress~\shortcite{Liu2020Autocompress} & 94.1\% & 94.0\% & 18.5$\times$ & Structured & 3.50/3.36\\
 \textbf{SS-Auto} & 94.1\% & 94.2\% & 24.6$\times$ & Structured & 3.39/3.28\\
 \hline
 \hline
 \multicolumn{6}{|c|}{MobileNet-V2} \\
 \hline
 DCP~\shortcite{Zhuang2018DCP} & 94.5\% & 94.7\% & 1.4$\times$ & Structured & N/A \\
 \textbf{SS-Auto} & 94.5\% & 94.6\% & 7.5$\times$ & Structured & 2.98/2.97 \\
 
 \hline
\end{tabular}
}\vspace{-4.0ex}
\end{table}

\begin{table}[!t]\scriptsize
\vspace{13pt}
\caption{\textbf{SS-Auto} vs. other pruning methods on CIFAR-100.} \label{Table2}
\resizebox{0.48 \textwidth}{!}{
\label{table1}
\begin{tabular}{|cccccc|}
\hline
 Methods & \begin{tabular}[c]{@{}c@{}}Base \\ Accuracy\end{tabular} & \begin{tabular}[c]{@{}c@{}}Prune\\ Accuracy\end{tabular} & \begin{tabular}[c]{@{}c@{}}Conv\\ Comp.rate\end{tabular} & \begin{tabular}[c]{@{}c@{}}Sparsity\\ Type\end{tabular} &
 \begin{tabular}[c]{@{}c@{}}Inference\\ Time (ms) \\CPU/GPU\end{tabular}\\ 
 \hline
 \hline
 \multicolumn{6}{|c|}{VGG-16} \\
 \hline
 Deco.~\shortcite{zhu2018ijcai} & 73.1\% & 73.2\% & 3.9$\times$ & Structured & N/A \\
 \textbf{SS-Auto} & 72.9\% & 72.2\% & 17.5$\times$ & Structured & 2.98/3.07\\ 
 \hline
 \hline
 \multicolumn{6}{|c|}{ResNet-18} \\
 \hline
 N2N~\shortcite{Ashok2017N2NLN} & 72.2\% & 68.0\% & 4.64$\times$ & Irregular & N/A\\
 \textbf{SS-Auto} & 76.0\% & 75.9\% & 5.9$\times$ & Structured & 4.14/3.82\\
 \textbf{SS-Auto} & 76.0\% & 75.3\% & 8.4$\times$ & Structured & 3.91/ 3.70\\
 \hline
 \hline
 \multicolumn{6}{|c|}{MobileNet-V2} \\
 \hline
 
 \textbf{SS-Auto} & 78.6\% & 78.7\% & 3.8$\times$ & Structured & 3.63/3.50 \\
 
 \hline
\end{tabular}
}
\vspace{-2.0ex}
\end{table}

\subsection{Results and Discussions on CIFAR-10 Dataset}
Table \ref{Table1} shows the comparison results on CIFAR-10 dataset. To measure the performance of the searched pruning rates fairly, we implement AutoCompress without purification. With the joint row and column pruning, as well as the automatically determined pruning rates, better pruning results are received. From the table, we can see that SS-Auto reaches the highest compression rate while preserving the accuracy compared with other pruning methods, which prove the effectiveness of SS-Auto in finding sparse structures. More specifically, SS-Auto achieves 34.4$\times$ compression rate with no accuracy loss and 41.1$\times$ compression rate with minor accuracy loss on VGG-16 with CIFAR-10 dataset. As for ResNet-18 and MobileNet-V2, we reach 24.6 $\times$ and 7.5$\times$ compression rate without accuracy degradation, respectively. 

The results also demonstrate that the per-layer pruning rates derived automatically by SS-Auto are reasonable and appropriate. One reason for the superiority of SS-Auto over the ADMM-based solution is that the manually determined pruning rates used in ADMM typically suffer from sub-optimality. With a much shorter time in finding the pruning rates, SS-Auto can reach similar or even better pruning results compared with AutoCompress. Figure~\ref{fig:weght in layer} shows the comparison of SS-Auto with AutoCompress on the counts of remaining parameters in each CONV layer of VGG-16 on CIFAR-10 dataset. As depicted in the figure, SS-Auto retains more weights in the early layers, which might help preserve the feature extraction in early stages for accuracy concerns, while removes more weights in the latter layers compared with AutoCompree. Moreover, the measured inference speedup on mobile CPU/GPU further validates the effectiveness of our proposed SS-Auto framework.

\subsection{Results and Discussions on CIFAR-100 Dataset}
The comparison results on CIFAR-100 dataset are demonstrated in Table \ref{Table2}. SS-Auto achieves 4.5$\times$ improvement in weight reduction compared with Decorrelation on VGG-16 and 1.8$\times$ improvement compared with N2N Learning on ResNet-18, under the same (or higher for SS-Auto) accuracy. The performance improvement is attributed to the joint row pruning and column pruning and the automatically determined per-layer pruning rate. By searching the row-wise and column-wise sparsity simultaneously, there is more flexibility in pruning, providing a higher probability of finding a better sparse structure. Furthermore, we notice that the GPU inference time of all pruned models is only slightly shorter than the CPU inference time. We ascribe this situation to the high compression rates reached by our proposed SS-Auto framework. After pruning, the layers shrink too much to exploit the parallelism of GPUs fully.

%% file: main.bbl
\begin{thebibliography}{}

\bibitem[\protect\citeauthoryear{Ashok \bgroup \em et al.\egroup
  }{2018}]{Ashok2017N2NLN}
Anubhav Ashok, Nicholas Rhinehart, Fares Beainy, et~al.
\newblock N2n learning: Network to network compression via policy gradient
  reinforcement learning.
\newblock In {\em IJCAI}, 2018.

\bibitem[\protect\citeauthoryear{Boyd \bgroup \em et al.\egroup
  }{2011}]{boyd2011distributed}
Stephen Boyd, Neal Parikh, Eric Chu, Borja Peleato, and Jonathan Eckstein.
\newblock Distributed optimization and statistical learning via the alternating
  direction method of multipliers.
\newblock {\em Foundations and Trends{\textregistered} in Machine Learning},
  2011.

\bibitem[\protect\citeauthoryear{Guo \bgroup \em et al.\egroup
  }{2016}]{guo2016dynamic}
Yiwen Guo, Anbang Yao, and Yurong Chen.
\newblock Dynamic network surgery for efficient dnns.
\newblock In {\em NeurIPS}, 2016.

\bibitem[\protect\citeauthoryear{Han \bgroup \em et al.\egroup
  }{2015}]{han2015learning}
Song Han, Jeff Pool, John Tran, and William Dally.
\newblock Learning both weights and connections for efficient neural network.
\newblock In {\em NeurIPS}, 2015.

\bibitem[\protect\citeauthoryear{Han \bgroup \em et al.\egroup
  }{2016}]{han2016eie}
Song Han, Xingyu Liu, Huizi Mao, Jing Pu, Ardavan Pedram, Mark~A Horowitz, and
  William~J Dally.
\newblock Eie: efficient inference engine on compressed deep neural network.
\newblock In {\em ISCA}, 2016.

\bibitem[\protect\citeauthoryear{He \bgroup \em et al.\egroup
  }{2016}]{he2016deep}
Kaiming He, Xiangyu Zhang, Shaoqing Ren, and Jian Sun.
\newblock Deep residual learning for image recognition.
\newblock In {\em CVPR}, 2016.

\bibitem[\protect\citeauthoryear{He \bgroup \em et al.\egroup
  }{2018}]{he2018amc}
Yihui He, Ji~Lin, Zhijian Liu, Hanrui Wang, Li-Jia Li, and Song Han.
\newblock Amc: Automl for model compression and acceleration on mobile devices.
\newblock In {\em ECCV}, 2018.

\bibitem[\protect\citeauthoryear{Hinton \bgroup \em et al.\egroup
  }{2012}]{hinton2012deep}
Geoffrey Hinton, Li~Deng, Dong Yu, George~E Dahl, Abdel-rahman Mohamed, et~al.
\newblock Deep neural networks for acoustic modeling in speech recognition: The
  shared views of four research groups.
\newblock {\em IEEE Signal Processing Magazine}, 2012.

\bibitem[\protect\citeauthoryear{Hong \bgroup \em et al.\egroup
  }{2016}]{hong2016convergence}
Mingyi Hong, Zhi-Quan Luo, and Meisam Razaviyayn.
\newblock Convergence analysis of alternating direction method of multipliers
  for a family of nonconvex problems.
\newblock {\em SIAM Journal on Optimization}, 2016.

\bibitem[\protect\citeauthoryear{Krizhevsky \bgroup \em et al.\egroup
  }{2012}]{krizhevsky2012imagenet}
Alex Krizhevsky, Ilya Sutskever, and Geoffrey~E Hinton.
\newblock Imagenet classification with deep convolutional neural networks.
\newblock In {\em NeurIPS}, 2012.

\bibitem[\protect\citeauthoryear{Li \bgroup \em et al.\egroup
  }{2016}]{li2016pruning}
Hao Li, Asim Kadav, Igor Durdanovic, Hanan Samet, and Hans~Peter Graf.
\newblock Pruning filters for efficient convnets.
\newblock {\em arXiv preprint arXiv:1608.08710}, 2016.

\bibitem[\protect\citeauthoryear{Liu \bgroup \em et al.\egroup
  }{2018}]{liu2018rethinking}
Zhuang Liu, Mingjie Sun, Tinghui Zhou, Gao Huang, and Trevor Darrell.
\newblock Rethinking the value of network pruning.
\newblock {\em arXiv preprint arXiv:1810.05270}, 2018.

\bibitem[\protect\citeauthoryear{Liu \bgroup \em et al.\egroup
  }{2020}]{Liu2020Autocompress}
Ning Liu, Xiaolong Ma, Zhiyuan Xu, Yanzhi Wang, Jian Tang, and Jieping Ye.
\newblock Autocompress: An automatic dnn structured pruning framework for
  ultra-high compression rates.
\newblock In {\em AAAI}, 2020.

\bibitem[\protect\citeauthoryear{Ma \bgroup \em et al.\egroup
  }{2020}]{ma2019tiny}
Xiaolong Ma, Geng Yuan, Sheng Lin, Caiwen Ding, Fuxun Yu, Tao Liu, Wujie Wen,
  Xiang Chen, and Yanzhi Wang.
\newblock Tiny but accurate: A pruned, quantized and optimized memristor
  crossbar framework for ultra efficient dnn implementation.
\newblock In {\em ASP-DAC}, 2020.

\bibitem[\protect\citeauthoryear{Min \bgroup \em et al.\egroup
  }{2018}]{min20182pfpce}
Chuhan Min, Aosen Wang, Yiran Chen, Wenyao Xu, and Xin Chen.
\newblock 2pfpce: Two-phase filter pruning based on conditional entropy.
\newblock {\em arXiv preprint arXiv:1809.02220}, 2018.

\bibitem[\protect\citeauthoryear{Parikh and Boyd}{2014}]{parikh2014proximal}
Neal Parikh and Stephen Boyd.
\newblock Proximal algorithms.
\newblock {\em Foundations and Trends{\textregistered} in Optimization}, 2014.

\bibitem[\protect\citeauthoryear{Real \bgroup \em et al.\egroup
  }{2017}]{real2017ICML}
Esteban Real, Sherry Moore, Andrew Selle, Saurabh Saxena, Yutaka~Leon Suematsu,
  Jie Tan, Quoc~V. Le, and Alexey Kurakin.
\newblock Large-scale evolution of image classifiers.
\newblock In {\em ICML}, 2017.

\bibitem[\protect\citeauthoryear{Ren \bgroup \em et al.\egroup
  }{2019}]{ren2019ADMMNN}
Ao~Ren, Tianyun Zhang, Shaokai Ye, Wenyao Xu, Xuehai Qian, Xue Lin, and Yanzhi
  Wang.
\newblock Admm-nn: an algorithm-hardware co-design framework of dnns using
  alternating direction methods of multipliers.
\newblock In {\em ASPLOS}, 2019.

\bibitem[\protect\citeauthoryear{Simonyan and
  Zisserman}{2014}]{simonyan2014very}
Karen Simonyan and Andrew Zisserman.
\newblock Very deep convolutional networks for large-scale image recognition.
\newblock {\em arXiv preprint arXiv:1409.1556}, 2014.

\bibitem[\protect\citeauthoryear{Wang \bgroup \em et al.\egroup
  }{2019}]{Wang2019NonstructuredDW}
Yanzhi Wang, Shaokai Ye, Zhezhi He, Xiaolong Ma, et~al.
\newblock Non-structured dnn weight pruning considered harmful.
\newblock {\em arXiv preprint arXiv:1907.02124}, 2019.

\bibitem[\protect\citeauthoryear{Wen \bgroup \em et al.\egroup
  }{2016}]{wen2016learning}
Wei Wen, Chunpeng Wu, , Yandan Wang, Yiran Chen, and Hai Li.
\newblock Learning structured sparsity in deep neural networks.
\newblock In {\em NeurIPS}, 2016.

\bibitem[\protect\citeauthoryear{Wu \bgroup \em et al.\egroup
  }{2016}]{Wu2016GooglesNM}
Yonghui Wu, Mike Schuster, Zhifeng Chen, Quoc~V. Le, Mohammad Norouzi, et~al.
\newblock Google's neural machine translation system: Bridging the gap between
  human and machine translation.
\newblock {\em arXiv preprint arXiv:1609.08144}, 2016.

\bibitem[\protect\citeauthoryear{Zhang \bgroup \em et al.\egroup
  }{2018a}]{zhang2018systematic}
Tianyun Zhang, Shaokai Ye, Kaiqi Zhang, Jian Tang, Wujie Wen, Makan Fardad, and
  Yanzhi Wang.
\newblock Systematic weight pruning of dnns using alternating direction method
  of multipliers.
\newblock In {\em ECCV}, 2018.

\bibitem[\protect\citeauthoryear{Zhang \bgroup \em et al.\egroup
  }{2018b}]{zhang2018adam}
Tianyun Zhang, Kaiqi Zhang, Shaokai Ye, Jian Tang, Wujie Wen, Xue Lin, Makan
  Fardad, and Yanzhi Wang.
\newblock Adam-admm: A unified, systematic framework of structured weight
  pruning for dnns.
\newblock {\em arXiv preprint arXiv:1807.11091}, 2018.

\bibitem[\protect\citeauthoryear{Zhao \bgroup \em et al.\egroup
  }{2019}]{Zhao2019VariationalCN}
Chenglong Zhao, Bingbing Ni, Jian Zhang, Qiwei Zhao, Wenjun Zhang, and Qi~Tian.
\newblock Variational convolutional neural network pruning.
\newblock In {\em CVPR}, 2019.

\bibitem[\protect\citeauthoryear{Zhu \bgroup \em et al.\egroup
  }{2018}]{zhu2018ijcai}
Xiaotian Zhu, Wengang Zhou, and Houqiang Li.
\newblock Improving deep neural network sparsity through decorrelation
  regularization.
\newblock In {\em IJCAI}, 2018.

\bibitem[\protect\citeauthoryear{Zhuang \bgroup \em et al.\egroup
  }{2018}]{Zhuang2018DCP}
Zhuangwei Zhuang, Mingkui Tan, Bohan Zhang, et~al.
\newblock Discrimination-aware channel pruning for deep neural networks.
\newblock In {\em NeurIPS}, 2018.

\bibitem[\protect\citeauthoryear{Zoph and Le}{2017}]{zoph2017NAS}
Barret Zoph and Quoc~V. Le.
\newblock Neural architecture search with reinforcement learning.
\newblock In {\em ICLR}, 2017.

\end{thebibliography}
